\documentclass[11pt,a4paper]{article}
\usepackage{times,latexsym}
\usepackage{url}
\usepackage[T1]{fontenc}

\usepackage[acceptedWithA]{tacl2021v1}
\hypersetup{
    colorlinks=true,
    linkcolor=black,
    filecolor=magenta,      
    urlcolor=black!100,
    }

\usepackage{tacl2021v1}

\usepackage{xspace,mfirstuc,tabulary}

\newif\iftaclinstructions
\taclinstructionsfalse 
\iftaclinstructions
\renewcommand{\confidential}{}
\renewcommand{\anonsubtext}{(No author info supplied here, for consistency with
TACL-submission anonymization requirements)}
\newcommand{\instr}
\fi

\iftaclpubformat 

\else

\fi

\title{State of What Art?\\ A Call for Multi-Prompt LLM Evaluation}

\author{
  Moran Mizrahi$^\dagger$ 
  \;
  Guy Kaplan$^\dagger$
  \;
  Dan Malkin$^\dagger$
  \;
  Rotem Dror$^\diamond$
  \;
  Dafna Shahaf$^\dagger$
  \;
  Gabriel Stanovsky$^\dagger$
  \\
  \ \\
  $^\dagger$School of Computer Science, The Hebrew University of Jerusalem
  \\
  $^\diamond$Department of Information Systems, University of Haifa
  \\
  \small{\texttt{\{moran.mizrahi, guy.kaplan2, dan.malkinhueb, gabriel.stanovsky\}@mail.huji.ac.il}}
  \\
  \small{\texttt{rdror@is.haifa.ac.il, dshahaf@cs.huji.ac.il}}
}
   
\date{}

\usepackage{graphicx}
\usepackage{booktabs}
\usepackage{multirow}
\usepackage{amsmath}
\usepackage[normalem]{ulem}
\usepackage{caption}
\usepackage{subcaption}
\usepackage{mathtools}
\usepackage{amsfonts}
\usepackage{longtable}
\usepackage{supertabular}
\usepackage{hyperref}

\usepackage[normalem]{ulem}
\usepackage{booktabs}
\usepackage{multirow}
\usepackage{cleveref}
\usepackage{changepage}

\definecolor{blue-violet}{rgb}{0.54, 0.17, 0.89}
\definecolor{blue-violet2}{rgb}{0.9, 0.35, 0.89}
\definecolor{shamrockgreen}{rgb}{0.0, 0.62, 0.38}
\definecolor{psychedelicpurple}{rgb}{0.7, 0.0, 1.0}
\definecolor{azure}{rgb}{0.0, 0.5, 1.0}

\usepackage{tikz}
\usepackage{calc}
\def\checkmark{\tikz\fill[scale=0.4](0,.35) -- (.25,0) -- (1,.7) -- (.25,.15) -- cycle;} 

\usepackage{longtable}
\usepackage{booktabs}

\newcommand{\remove}[1]{}

\newcommand{\xhdrnodot}[1]

\newcommand{\draftcomment}[3]{{\textcolor{#3}{[#1 -- #2]}}}

\newcommand{\mnote}[1]{\draftcomment{#1}{\textsc{moran}}{blue-violet2}} 

\newcommand{\rotem}[1]{\draftcomment{#1}{\textsc{rotem}}{orange}}  

\newcommand{\gabi}[1]{\draftcomment{#1}{\textsc{gabi}}{red}} 
\newcommand{\gabis}[1]{\gabi{#1}}

\newcommand{\dan}[1]{\draftcomment{#1}{\textsc{dan}}{shamrockgreen}}

\newcommand{\bbl}[0]{\textsc{BBL}}
\newcommand{\bbh}[0]{\textsc{BBH}}

\newcommand{\lmentry}[0]{\textsc{LMentry}}

\newcommand{\bigbench}[0]{BIG-bench}
\newcommand{\bigbenchlite}[0]{BIG-bench Lite}
\newcommand{\bigbenchhard}[0]{BIG-bench Hard}

\newcommand{\gptturbo}[0]{GPT-3.5-Turbo}

\newcommand{\cgpt}[0]{\gptturbo{}}
\newcommand{\llama}[0]{LLaMA}

\newcommand{\flan}[0]{Flan-T5}

\newcommand{\flanlarge}[0]{Flan-T5-large}
\newcommand{\flanxl}[0]{Flan-T5-XL}

\newcommand{\alpaca}[0]{Alpaca}

\newcommand{\falcon}[0]{Falcon-Instruct-7b}

\newcommand{\tzero}[0]{T0}
\newcommand{\tzeropp}[0]{T0pp}

\newcommand{\vicunawp}[0]{Vicuna-13b}
\newcommand{\ultralm}[0]{UltraLM}
\newcommand{\airoboros}[0]{Airoboros}
\newcommand{\noushermes}[0]{Nous-Hermes}
\newcommand{\minotaur}[0]{Minotaur}
\newcommand{\numofmodelsopen}[0]{16}
\newcommand{\numofmodelsall}[0]{20}
\newcommand{\numoftasks}[0]{39}
\newcommand{\numofsamples}[0]{6.5M}

\newcommand{\numoftaskswithnegkendalltaut}[0]{15}
\newcommand{\openai}[0]{OpenAI}

\makeatletter
\newcommand\footnoteref[1]{\protected@xdef\@thefnmark{\ref{#1}}\@footnotemark}
\makeatother

\begin{document}
\maketitle

\begin{abstract}
Recent advances in LLMs have led to an abundance of evaluation benchmarks, which typically rely on a \emph{single instruction template} per task. We create a large-scale collection of instruction paraphrases and comprehensively analyze the brittleness introduced by single-prompt evaluations across 6.5M instances, involving 20 different LLMs and 39 tasks from 3 benchmarks. We find that different instruction templates lead to very different performance, both absolute and relative. Instead, we propose a set of diverse metrics on \textit{multiple instruction paraphrases}, specifically tailored for different use cases (e.g., LLM vs.~downstream development), ensuring a more reliable and meaningful assessment of LLM capabilities. We show that our metrics provide new insights into the strengths and limitations of current LLMs.


\end{abstract}

\section{Introduction}
\label{sec:intro}


Recent years have seen an explosion of large language models (LLMs), which generalize to unseen tasks via natural language instructions. Various LLM evaluation benchmarks, such as \bigbench{} and HELM, use a \emph{single} instruction template per task,  evaluating all models against it~\citep{srivastava2022beyond,liang2022holistic}. 
However, there could be a myriad of ways to phrase an instruction template for a given task; see  Figure~\ref{fig:main} for examples of different templates for the task of recognizing homophones. Naturally, LLM performance depends on the chosen template.


We explore the question of \emph{robustly comparing different models on a given task}. We first create a dataset of paraphrased instructions, employing three automatic paraphrasing methods based on recent techniques such as chain-of-thought. We manually verify and filter a large collection of more than 175 paraphrases for different tasks (5K instruction paraphrases in total), which we make publicly available for future research.\footnote{\href{https://github.com/SLAB-NLP/Multi-Prompt-LLM-Evaluation}{github.com/SLAB-NLP/Multi-Prompt-LLM-Evaluation}}

Next, we use our dataset to perform a large-scale statistical evaluation of over \numofsamples{} instances, involving \numofmodelsall{} different LLMs and \numoftasks{} tasks from 3 benchmarks. 
We find that models perform very differently on different instruction paraphrases. 
For example, Figure~\ref{fig:main} shows four models evaluated on four semantically equivalent prompts, with both absolute and relative performance varying widely;
%
%
%
one can even observe cases where the same model performs \emph{the best} on one instruction and \emph{the worst} on a semantically equivalent instruction (e.g., \gptturbo{} on $P_1$ vs.~$P_4$). 
%
%
Subsequently, we argue that \emph{very little can be said} on either absolute or relative performance based on single-instruction evaluation. This may also partially explain why some models seem less accurate in practice than their formal evaluation suggests.

Note that while the claim that evaluating against a single instruction template leads to brittle results is not surprising per se, to the best of our knowledge it has never been subjected to rigorous empirical testing before.

To address the limitations of single-instruction evaluation, we propose to take a step back and consider {\emph{multi-prompt LLM evaluation} --- a set of metrics which measure aggregated performance over a set of instruction template paraphrases}. 

We argue that different use cases should entail different evaluation metrics. 
For example, LLM developers may be interested in measuring the \emph{robustness of performance} across multiple instruction templates. 
In contrast, {developers aiming to integrate an LLM into a specific} downstream task may be {interested in comparing  models}  according to their corresponding \emph{top-performing} instruction. 



We evaluate \numofmodelsall{} LLMs with our metrics, finding that their absolute and relative performance differ from results obtained with the benchmarks’ original instructions. 
We demonstrate that different models excel in different metrics: 
For instance, in the \lmentry{} benchmark, \llama{}-based models are comparable to T5-based models when looking at top-performing instructions, but lag behind when average performance is considered, due to poor performance on a large number of paraphrases.
We also show that our automatic paraphrasing method is effective, and there is no need to manually verify the paraphrases.

Our results suggest that future work should {use multi-prompt LLM evaluations and choose a metric for aggregating the results according to the \emph{extrinsic needs} of the evaluators. We hope that our work will help spur more consistency and comparability in LLM evaluation, which is strongly tied to real-world usage of LLMs.

\begin{figure}[t!]
\includegraphics[width=\linewidth]{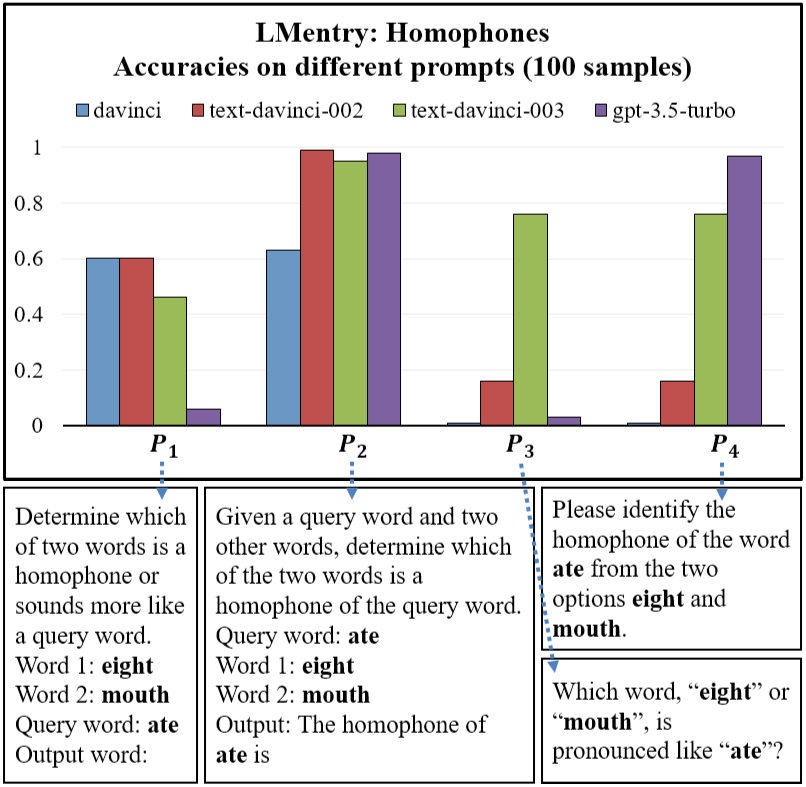}
\caption{\label{fig:main}
Evaluation of different OpenAI models on the homophones task from \lmentry{} over four paraphrases. Each cluster of columns corresponds to a distinct paraphrased \emph{instruction template} (see respective texts below; words in bold indicate an instantiation). Despite all instructions being semantically equivalent, both absolute performance and relative ranking vary widely.}\end{figure}

\section{Background and Definitions} 
\label{sec:background_and_definitions}

Below we survey how generalization to a new task format is evaluated and compared between LLMs, finding that the common practice involves a single (or very few) task instruction templates.
In the rest of the paper, we will argue that such practice leads to brittle, unreliable results. 

\paragraph{Task instruction templates.}
\label{sec:inst_temp_def}
Following \citet{mishra2021cross, chung2022scaling}, we separate between task instruction, samples, and input-output exemplars which may be provided during in-context learning.
We define an \emph{instruction template} for a given task as a string with placeholders where the input samples are to be inserted. 
As seen in Figure~\ref{fig:main}, the same task can be described using different task instruction templates. 



\paragraph{Evaluation benchmarks.}
Several recent efforts aim to standardize LLM evaluation.
Notable examples include MMLU~\citep{hendrycks2020measuring}, \bigbench{}~\citep{srivastava2022beyond,suzgun2022challenging}, and HELM~\citep{liang2022holistic}. In all of these, each task has a single instruction template, against which all models are evaluated. 
Another benchmark, \lmentry{}~\citep{efrat2022lmentry}, reports models' average performance on three instruction templates.
The instruction templates are provided with these benchmarks, allowing new models to be tested against the same template. 

{We note that many notable works do not disclose the}  instruction templates 
{used for evaluation} (e.g., LLaMA \citep{touvron2023llama}, PALM~\citep{chowdhery2022palm}, GPT-4~\citep{Achiam2023GPT4TR}, Gemini~\citep{Anil2023GeminiAF}). {While there are reasons to withhold instructions (e.g., avoid potential leakage),} this {practice} exacerbates the challenge of meaningful comparative evaluation. 


\paragraph{Prompt robustness.}
Related to this study is a line of work measuring LLM's robustness to prompt (or instruction template) modifications. Unlike our work, these typically
aim to measure model performance against \emph{adversarial} paraphrasing approaches.
PromptBench~\citep{zhu2023promptbench} measures performance on erroneous instructions (e.g.,  instructions written by non-native English speakers). They then compare performance on perturbed instructions vs.~the benchmark's original instructions, which are considered the gold-standard reference. 
\citet{gu2022robustness} examined~a single LLM's robustness under various instruction perturbations, including word-, sentence-, and instruction-level changes. 
\citet{sun2023evaluating} show that LLMs perform better on instructions they have seen in training 
compared to manual paraphrases.
We later incorporate their manual paraphrases in our evaluation of \bigbenchlite{}.

In contrast to works on prompt robustness, we analyze the 
impact of the choice of prompt in terms of both absolute and relative model performance, covering a wide range of models and several different metrics.

\section{Experimental Setup}
\label{sec:background}

\subsection{Tasks}
\label{sec:method-bmks}


We evaluate \numoftasks{} diverse tasks from three evaluation benchmarks, as itemized below. 

\paragraph{10 tasks from \lmentry{}~\cite{efrat2022lmentry}.} \lmentry{} consists of simple linguistic tasks (e.g., ``write a word that doesn't contain the letter $l$''), {each accompanied by three associated instruction templates.}
The tasks are designed to capture explainable and controllable linguistic phenomena. We choose {the} 10 tasks that received the lowest scores in the original paper, as these more challenging tasks are likely to better highlight the differences between models.

\paragraph{14 tasks from \bigbenchlite{} (\bbl{};~\citealp{srivastava2022beyond}).} These cover multiple knowledge domains, sampled from the larger BIG-Bench benchmark~\citep{srivastava2023beyond}.
We focus on a set of 14 tasks studied recently by \citet{sun2023evaluating}. Each task in \bbl{} is associated with a single instruction template.
 

\paragraph{15 tasks from \bigbenchhard{} (\bbh{}; \citealp{suzgun2022challenging}).} This is another curated subset of \bigbench{}, containing particularly challenging tasks on which LLM underperform the average human score. We focused on a set of 15 classification and multiple choice tasks to streamline the evaluation process. Each task in \bbh{} is associated with a single instruction template.

\begin{table}[tb!]
\small
\centering
\resizebox{\columnwidth}{!}{%
\begin{tabular}{@{}lllr@{}}
\toprule
\textbf{Model}                                            &  \textbf{Model size}  & \textbf{Base model} & \textbf{\# Params}\\ \midrule
\multirow{5}{*}{\flan{}}   & Small   & \multirow{5}{*}{T5
} &  80M    \\
                                                    & Base   &&  250M    \\
                                                    & Large  & &  780M    \\
                                                    & XL     & &  3B    \\
                                                    & XXL    & &  11B    \\ 
                                                    \midrule
 \multirow{2}{*}{\tzero{}} & Small   &  \multirow{2}{*}{T5 
}  & 3B   \\
                                                    & \tzeropp{} &   & 11B     \\ \midrule
\multirow{2}{*}{\alpaca{}}  & Small & \multirow{2}{*}{LLaMA 
} & 7B     \\
                                                    & Big   && 13B      \\ \midrule

Vicuna                  &       &     LLaMA 
& 13B      \\ 
\airoboros{}              &             &  LLaMA 
& 13B     \\ 
\ultralm{}              &         &      LLaMA 
& 13B     \\ 
\noushermes{}            &          &     LLaMA 
& 13B     \\ 

 Falcon-Instruct                  &  & Falcon 
& 7B   \\ 
MPT                 &   &    MPT 
& 7B   \\ 

\minotaur{}               &            &   StarCoder Plus 
& 15B     \\ 
                                                    \bottomrule
\end{tabular}%
}
\caption{The different LLMs evaluated in this work, grouped by model family, along with their size, in number of parameters. All models were instruction-tuned.}
\label{tab:llms}
\end{table}

\paragraph{Measuring performance.}
In \lmentry{} we measure performance using the official evaluation script, while in Big-Bench we perform exact string matching. 
We note that while exact matching is somewhat strict, we believe  it is also fair and straightforward.





\remove{
Furthermore, Given that each task involves a substantial number of instruction templates (ranging from 130 to 260), we opt to conserve computational resources by evaluating each instruction template on a randomly selected subset of 100 samples. 
Table 1 provides a comprehensive overview, specifying the number of prompt paraphrases available for each task, the method used to prompt models on that task (zero-shot or one-shot), and the sample size employed in assessing each paraphrase for each task.
}

\subsection{Models}
\label{sec: models}

We evaluate \numofmodelsopen{} {instruction-tuned} LLMs from 11 diverse model families \citep{chung2022scaling,sanh2021multitask, taori2023alpaca, zheng2023judging, Durbin2023, ding2023enhancing, NousResearch2023, almazrouei2023falcon, MosaicML2023Introducing, collective2023} (see Table~\ref{tab:llms}).
We refrain from including closed, API-based models (e.g., \openai{} models) in our main evaluation for two reasons. First, using them at scale is an expensive prospect. For example, running our entire evaluation suite on GPT-4 will cost thousands of dollars. Second, and more importantly, the closed API for these models reportedly manipulates the input prompts in an undisclosed manner (e.g., wrapping them with meta-prompts, or rerouting to other models)~\citep{Rao2023TrickingLI} which interferes with our evaluation. We do however perform a small-scale evaluation of \openai{} models in Section~\ref{sec:openai_evaluation} to show that they are also sensitive to prompt paraphrasing.


\section{{Evaluating against a Single Prompt} Leads to {Instability in} Results}
\label{sec:flaws_in_cur_eval_methods}

\remove{
\begin{figure}[t!]
\includegraphics[width=\textwidth]{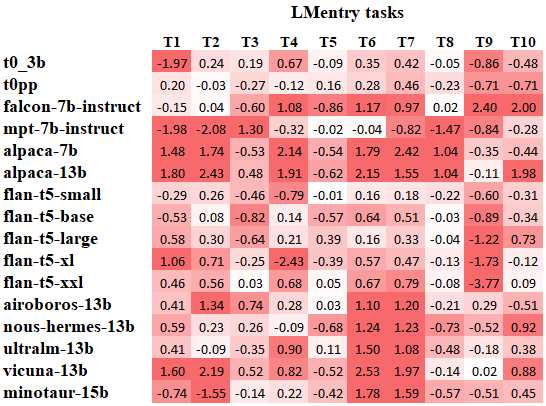}
\caption{\label{fig:divergence_lmentry}
{Model and task performance divergence. For each task, this table shows the number of standard deviations by which the performance of each model on the original instruction templates deviates from the averaged model performance. Dark red cells indicate substantial divergence values exceeding one standard deviation. \gabi{if we leave these figures in the paper we need to improve them visually, e.g., using seaborn. Also, maybe merge with the other heatmap and put across two columns}}
}
\end{figure}

\begin{figure}[t!]
\includegraphics[width=\linewidth]{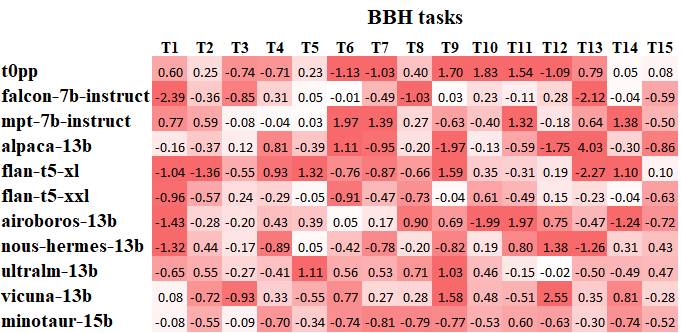}
\caption{\label{fig:divergence_bbh}
{Model and task performance divergence. For each task, this table shows the number of standard deviations by which the performace of each model on the original instruction templates deviates from the averaged model performance. Dark red cells indicate substantial divergence values exceeding one standard deviation.}
}\end{figure}
}



As discussed in the previous section, LLMs are usually evaluated against a single instruction template. 
In this section, we will show that this approach is quite brittle. Indeed, a simple rephrasing of the instruction template can lead to drastic changes in both absolute and relative model performance.

In Section~\ref{sec:prompt_generation_method} we create a large number of automatically-generated instruction paraphrases for { tasks from the \lmentry{} and \bbh{} benchmarks}. Paraphrases are created using an LLM and verified by human annotators. 
In Section~\ref{sec:brit_results}, we statistically analyze the performance of various LLMs against these instruction templates and quantify the variation in model performance. Finally, in Section~\ref{sec:manual}, we show that models 
exhibit similar brittleness with manually-written paraphrases for tasks from the \bbl{} benchmark.

\subsection{Paraphrasing Instruction Templates}
\label{sec:prompt_generation_method}

We use three prompting methods which were found useful in previous works: (1) instruction template rephrasing: asking an LLM to rephrase a seed prompt~\citep{lester2021power, gonen2022demystifying, honovich2022unnatural}; 
(2) Chain-of-Thought prompting~\citep{wei2022chain}: we provided the model with a sequence of steps in which the model is asked first to produce a task description, and then to generate various instruction templates for the task; 
and (3) Gradual template generation: inspired by \citet{honovich2022instruction}, we split the COT approach into three LLM calls. The first for generating a task description from a seed instruction template, the second for generating instruction provided by input-output examples, and the third for processing the instruction and examples into an instruction template. 

{In all of the above, we use \cgpt{} for generation, and} the original instruction templates for each of our tasks to seed these three generation methods, resulting on average in more than 200 automatically-generated instruction template paraphrases for each of our tasks (see Table~\ref{table:paraphrases_correctness_summary}).
We  make this collection, as well as the code used to generate it, publicly available for reproducibility and to enable future work.

\begin{table}[t!]
\resizebox{\columnwidth}{!}{%
\begin{tabular}{l|lccc}
\toprule
\textbf{Benchmark}          & \textbf{Method}                                                        & \textbf{\begin{tabular}[c]{@{}c@{}}\#Automatic \\ Paraphrases\end{tabular}}                                     & \textbf{\begin{tabular}[c]{@{}c@{}}\#Correct \\ Paraphrases\end{tabular}}                                  & \textbf{\begin{tabular}[c]{@{}l@{}}Correct \\ Ratio \end{tabular}}   \\                                               
\hline

\lmentry{}                           & \begin{tabular}[c]{@{}l@{}}All\\ Rephrase\\ CoT\\ Gradual\end{tabular} & \begin{tabular}[c]{@{}l@{}}2429\\ 461\\ 1286\\ 652\end{tabular} & \begin{tabular}[c]{@{}l@{}}2186\\ 408\\ 1234\\ 514\end{tabular} & \begin{tabular}[c]{@{}l@{}}90.00\%\\ 88.50\%\\ 95.96\%\\ 78.83\%\end{tabular}  \\ \midrule

\bbh{}                          & \begin{tabular}[c]{@{}l@{}}All\\ Rephrase\\ CoT\\ Gradual\end{tabular} & \begin{tabular}[c]{@{}l@{}}2615\\ 734\\ 775\\ 1091\end{tabular} & \begin{tabular}[c]{@{}l@{}}2209\\ 627\\ 630\\ 937\end{tabular}  & \begin{tabular}[c]{@{}l@{}}84.47\%\\ 85.42\%\\ 81.29\%\\ 85.88\%\end{tabular} \\
\bottomrule
\end{tabular}%
}
\caption{
{Manual validation and filtering of automatic instruction paraphrases generated for \lmentry{} and \bbh{}, showing percentages of valid paraphrases.}
}
\label{table:paraphrases_correctness_summary}
\end{table}
\paragraph{Manual validation and filtering of automatic instruction paraphrases.} 
All automatically generated paraphrases {were manually verified and filtered by an annotator from our group }
{to ensure their coherence and relevance to the task.} 
{A portion of the data involving 15 randomly selected templates from each task, totaling in 375 instructions, was also given to a second annotator;
%
} results  
show reliable agreement (Table~\ref{tab:agreement}), indicating  our evaluation process is calibrated.

See Table \ref{table:paraphrases_correctness_summary} for a fine-grained distribution across the different generation metrics. {Overall, we found that 90\% of the generated paraphrases created for \lmentry{} were correct, and roughly 84\% of the paraphrases for \bbh{} were correct.} 

On average, the validation process yields {240} validated instruction paraphrases per task {for} \lmentry{} and {175 paraphrases per task for} \bbh{}. {Next, we}
 use {these paraphrases} to quantify performance variability due to instruction template paraphrasing {across $\sim\numofsamples{}$ instances.}\footnote{Calculated as the number of models tested per task × number of paraphrased instructions per task × 100 samples, across all tasks and benchmarks $\approx 240\times16\times100\times10$ (\lmentry{}) $+ 175\times11\times100\times15$ (\bbh{}).}

\begin{table}[t!]
\centering
\small
\resizebox{\columnwidth}{!}{%
\begin{tabular}{@{}lccc@{}}
\toprule
\textbf{Benchmark} & \multicolumn{1}{l}{\textbf{Correct (\%)}} & \multicolumn{1}{l}{\textbf{\begin{tabular}[c]{@{}l@{}}Agreement\\ (accuracy)\end{tabular}}} & \multicolumn{1}{l}{\textbf{\begin{tabular}[c]{@{}l@{}}Agreement\\ (Cohen's $\kappa$)\end{tabular}} } \\ \midrule
\lmentry{}         & 86.0                                             & .953 & .774                                     \\
\bbh{}             & 86.7                                             & .916 & .491                                     \\ \bottomrule
\end{tabular}%
}
\caption{{Human evaluation of doubly annotated paraphrases. Out of 375 automatically generated instructions, more than 85\% were found to be correct by both annotators. Both Cohen's $\kappa$ and the agreement accuracy indicate varying, yet generally high levels of agreement given pronounced label imbalance.}
}
\label{tab:agreement}
\end{table}

\subsection{Quantifying Performance Variance due to Instruction Paraphrasing}
\label{sec:brit_results}
We leverage the collection of validated paraphrases to assess how model performance varies with paraphrasing.
%
Our main finding is that the common approach of evaluating against a single propmt is 
unstable, leading to unreliable results.

\paragraph{Instance sampling and prompt construction.}
Our study involves a large number of tasks, models, and instruction paraphrases. 
However, evaluating LLMs can become prohibitively expensive with the increase of the number of samples, datasets, models, and instruction templates~\citep{Perlitz2023EfficientB}. 
To make our evaluation feasible, we chose to evaluate each instruction template on a randomly selected subset of 100 task samples.
Furthermore, we found that all models struggle on \bbh{}, beyond the point of meaningful comparison. To address this, we evaluate  11 out of the 16 models on it (the ones with the largest number of parameters), and add an example of the prediction format to all instruction template paraphrases. 

Examining the effect of few-shot learning is beyond the scope of this paper, however,  \citet{Sclar2023QuantifyingLM}, \citet{weber2023mind} and \citet{voronov2024mind} recently observed similar performance sensibility when introducing varying number of in-context examples.



\begin{table}[t!]
\small
\resizebox{\columnwidth}{!}{%
\begin{tabular}{lcc}
\toprule
\textbf{Tasks} & \textbf{Kendall's W}  & \textbf{Friedman p-val}
\\ \hline
\textbf{\lmentry{}}  \\
 $\;$ not containing  & .271 (weak) & \textbf{0.0}{*}\\
 $\;$ word before  & .367 (weak) & \textbf{0.0}{*}\\
 $\;$ first alphabet  & .436 (weak) &\textbf{0.0}{*}\\
 $\;$ less letters  & .485 (weak) &\textbf{0.0}{*}\\
 $\;$ rhyming word  & .496 (weak) & \textbf{0.0}{*}\\
 $\;$ ends with word &  .518 (weak) & \textbf{0.0}{*}\\
 $\;$ homophones  & .518 (weak) &\textbf{0.0}{*}\\
 $\;$ all words  & .522 (weak) & \textbf{0.0}{*}\\
 $\;$ any words  & .527 (weak) & \textbf{0.0}{*}\\
 $\;$ more letters  & .540 (weak) & \textbf{0.0}{*}\\
  
\midrule

\textbf{\bigbenchhard{}} \\ 

$\;$ recommendations  &  .628 (medium) &.897\\
$\;$ formal fallacies  &   .704 (medium) &\textbf{5.6E-13}\\
$\;$ geometric shapes  & .710 (medium) &.167\\
$\;$ hyperbaton  &  .730 (medium) &\textbf{1.0E-4}\\
$\;$ logical deduction 3  &  .740 (medium) & \textbf{4.9E-16}\\
$\;$ disambiguation qa  &  .764 (medium) &\textbf{2.1E-17}\\
$\;$ ruin names  & .776 (medium) & .366\\
$\;$ logical deduction 7  & .778 (medium) & \textbf{1.4E-13}\\
$\;$ translation error  & .800 (medium) &\textbf{6.9E-9}\\
$\;$ logical deduction 5  & .818 (medium) &\textbf{3.0E-9}\\
$\;$ snarks  & .823 (medium) &.604\\
$\;$ penguins in a table  & .830 (medium) &\textbf{7.3E-15}\\
$\;$ navigate  &   .838 (medium) &\textbf{5.6E-10}\\
$\;$ causal judgement  & .851 (strong)& \textbf{4.9E-7}\\
$\;$ sports  & .873 (strong)&\textbf{8.0E-13}\\

\midrule
\textbf{\bigbenchlite{}} \\ 
$\;$ known unknown &  .316 (weak) & \textbf{4.4E-5}\\
$\;$ play dialog &  .355 (weak) & \textbf{4.3E-5}\\
$\;$ winowhy &   .520 (weak) & \textbf{6.0E-4}\\
$\;$ strategic qa &  .529 (weak) & .191\\
$\;$ hindu knowledge &   .560 (weak) & .569\\
$\;$ conceptual &   .731 (medium) & .132\\
$\;$ strange stories & .731 (medium) & .431\\
$\;$ code desc &   .756 (medium) & \textbf{.002}\\
$\;$ novel concepts &  .787 (medium) & .620\\
$\;$ logic grid puzzle &  .796 (medium) & \textbf{.010}\\
$\;$ lang. identification &   .811 (medium) & \textbf{.002}\\
$\;$ vitaminc &  .888 (strong) & .772\\
$\;$ bbq lite &  .890 (strong) & \textbf{.023}\\
$\;$ logical deduction &  .913 (strong) & .895\\
\bottomrule
\end{tabular}%
}
\caption{\label{tab:tasks_info2} 
{Kendall's $W \in [0,1]$ values for all tasks sorted in ascending order.} The smaller the value of $W$ the more that the ranking on different prompts is de-correlated. Most $W$ are smaller than $0.85$, indicating 
weak to moderate agreement.
The p-values from Friedman test indicate significant differences between rankings of models when using different prompts. $^*$p-values of 0 represent statistical significance levels that are smaller than 1E-50. 
} 
\end{table}

\remove{
\begin{table}[tb!]
\resizebox{\columnwidth}{!}{%
\begin{tabular}{lccc}
\toprule
\textbf{Benchmark \& Task} & \textbf{Kendall's W} & \textbf{Friedman p-val}
\\ \hline
\textbf{\lmentry{}}  \\
 $\;$ word not containing  & .271 &\textbf{0.0}{*}\\
 $\;$ word before  & .367 & \textbf{0.0}{*}\\
 $\;$ first alphabetically  & .436 &  \textbf{0.0}{*}\\
 $\;$ less letters  & .485 & \textbf{0.0}{*}\\
 $\;$ rhyming word  & .496 & \textbf{0.0}{*}\\
 $\;$ ends with word &  .518 &  \textbf{0.0}{*}\\
 $\;$ homophones  & .518 & \textbf{0.0}{*}\\
 $\;$ all words  & .522 &  \textbf{0.0}{*}\\
 $\;$ any words  & .527 &  \textbf{0.0}{*}\\
 $\;$ more letters  & .540 & \textbf{0.0}{*}\\
  
\midrule

\textbf{\bigbenchhard{}} \\ 

$\;$ movie recommendation  & .628 & .897\\
$\;$ formal fallacies  &   .704 & \textbf{5.6E-13}\\
$\;$ geometric shapes  & .710 & .167\\
$\;$ hyperbaton  &  .730 & \textbf{1.0E-4}\\
$\;$ logical deduction 3  &  .740 & \textbf{4.9E-16}\\
$\;$ disambiguation qa  &  .764 & \textbf{2.1E-17}\\
$\;$ ruin names  & .776 &  .366\\
$\;$ logical deduction 7  & .778 & \textbf{1.4E-13}\\
$\;$ salient translation error detection  & .800 & \textbf{6.9E-9}\\
$\;$ logical deduction five objects  & .818 & \textbf{3.0E-9}\\
$\;$ snarks  & .823 & .604\\
$\;$ penguins in a table  & .830 & \textbf{7.3E-15}\\
$\;$ navigate  &   .838 & \textbf{5.6E-10}\\
$\;$ causal judgement  & .851 & \textbf{4.9E-7}\\
$\;$ sports understanding  & .873 & \textbf{8.0E-13}\\

\bottomrule
\end{tabular}
}
\caption{\label{tab:tasks_info2} 
{Kendall's $W \in [0,1]$ values for all tasks sorted in ascending order.} The smaller the value of $W$ the more that the ranking on different prompts is de-correlated. Most $W$ are smaller than $0.85$, indicating less than optimal correlation. The p-values from Friedman test indicate significant differences between rankings of models when using different prompts. {In the table, p-values of 0.0 denoted with an asterisk symbol (*) represent statistical significance levels that are smaller than E-50. \gabis{maybe add W interpretation to the table (e.g., write ``moderate'' or ``weak'' agreement for each of thtasks?}}\mnote{\checkmark{}}
}
\end{table}
}

\begin{figure*}[t!]
\includegraphics[width=\textwidth]{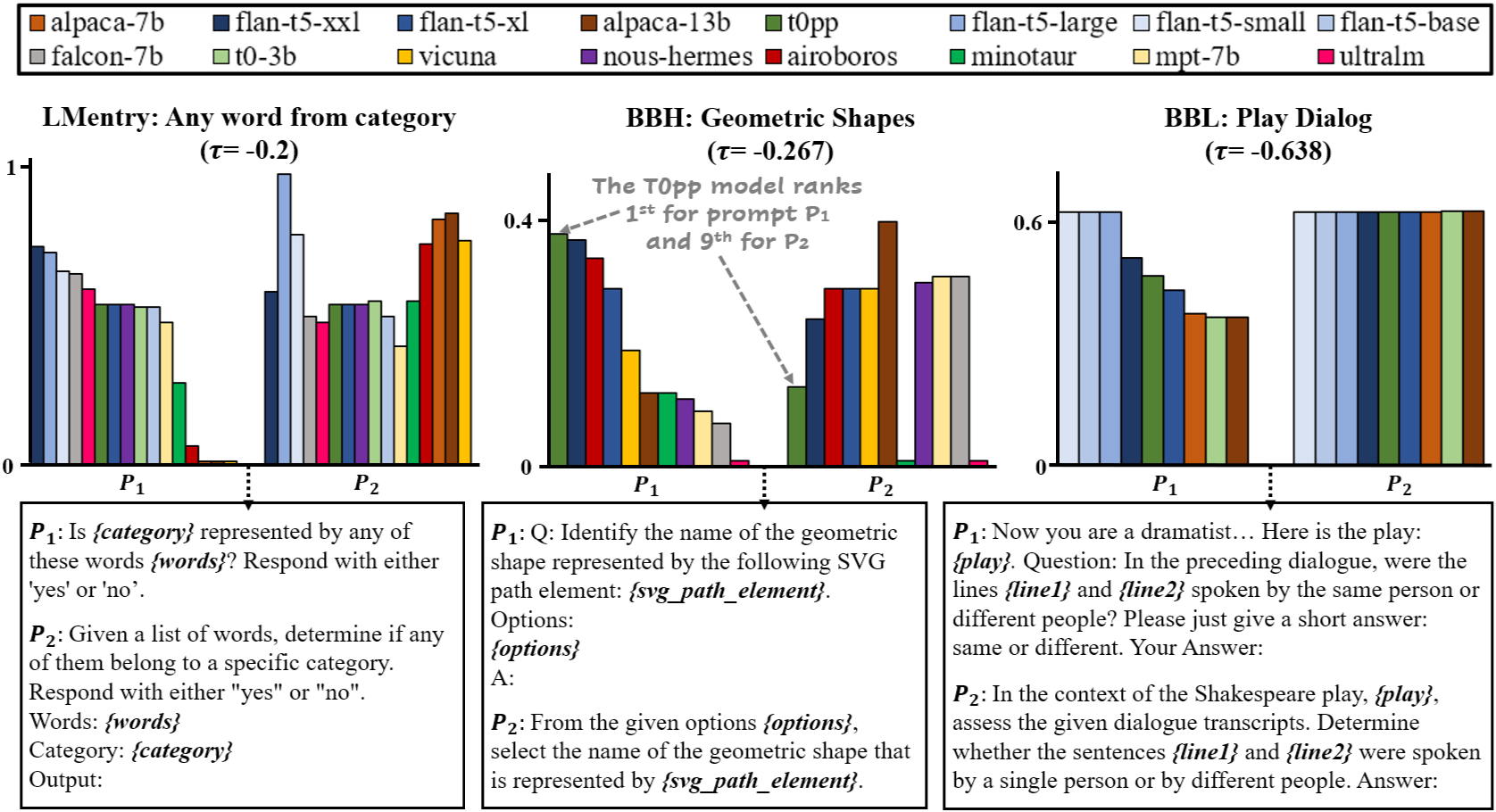}
\caption{\label{fig:tau-comparison} 
Model performance and ranking induced by pairs of paraphrases that exhibit the minimal Kendall $\tau$ correlation on three different tasks (one for each benchmark). For each template pair, models are ordered  according to their performance against the first instruction template $P_1$, enabling straightforward comparisons of ranking changes. In other words, if the bars of $P_2$ appear scattered rather than follow a clear descending order, this indicates a significant reshuffling of rankings. 
}
\end{figure*}
\paragraph{Using a single-instruction template leads to brittle ranking.} 
We compute Kendall’s $W: \mathbb{N}^{m \times n} \mapsto [0, 1]$~\citep{kendall1939problem}, a non-parametric statistic which measures the ranking correlation between $m$ judges (instruction templates, in our case) ranking $n$ objects (LLMs, in our case) by calculating the squared deviation between the sum of ranks of different judges ($R_i=\sum_{j=1}^{m}r_{ij}$) and their mean value:
\[W = \frac{12\sum_{i=1}^{n}(R_i-\bar{R})^2}{m^2(n^3-n)}\]

Kendall's $W$ would be $1$ for all tasks if model ranking were the same among all instruction templates (in other words, they are interchangeable for the sake of evaluation).  In contrast, the more $W$ approaches $0$,  the lesser the rankings induced by different instructions agree.

The results (Table~\ref{tab:tasks_info2}) demonstrate that a single instruction template leads to unreliable rankings for many of the tasks, with {10} of the tasks exhibiting only slight to moderate ranking agreement, and only two exhibiting strong agreement. To complement the analysis, we performed Friedman test with tied data~\citep{corder2011nonparametric}, showing that 
different instructions lead to statistically significant differences in performance for 21 out of the 25 tasks. 




\paragraph{Examples of differences in model ranking.} 
We illustrate the implications of ranking differences in Figure~\ref{fig:tau-comparison}.
In all three cases, $P_1$ and $P_2$ are valid paraphrases, yet they lead to vastly different rankings. For example, 
\tzeropp{} ranks first on the \bbh{} task (center) according to $P_1$ and only 9th according to $P_2$.
Similarly, \alpaca{}-13B and \alpaca{}-7B are in the \emph{top}-performing models on the \lmentry{} task $P_2$, while they rank \emph{last} for $P_1$. 

We quantify the difference between two rankings with Kendall's $\tau: \mathbb{N}^n \times \mathbb{N}^n \mapsto [-1,1]$, which estimates the agreement between two specific instruction templates which induce rankings $R_{1}, R_{2}$ over $n$ LLMs, formally defined as~\citep{kendall1945treatment}:
\begin{equation*}
    \tau_b = \frac{P-Q}{\sqrt{(P+Q+T)\cdot(P+Q+U)}}
\end{equation*}

Where $P$ is the number of concordant pairs, $Q$ is the number of discordant pairs, $T$ is the number of ties in the first ranking, and $U$ is the number of ties in the second ranking.
Therefore, $\tau > 0$ indicates that most pairs are concordant (with $\tau = 1$ indicating perfect agreement), and $\tau < 0$ indicates that most pairs are discordant (with $\tau = -1$ indicating perfect disagreement).
%
%
%
Overall, \numoftaskswithnegkendalltaut{} tasks out of 25 have instruction template paraphrases with negative Kendall's $\tau$, indicating mostly disagreeing LLM rankings.




\paragraph{Absolute model performance varies widely on single-instruction templates.}
Aside from vastly different relative model rankings, instruction template paraphrases often result in varying absolute model performances.
To quantify this variance, we calculated \textit{divergence}, defined as the number of standard deviations by which the performance, as assessed using the original instruction templates, deviates from the model’s average performance over all paraphrases.
%

The results in Figure~\ref{fig:divergence_lmentry_bbh} reveal noticeable divergence {for the \lmentry{} benchmark}, 
defined as surpassing one standard deviation~\citep{kazmier2003business}.
For instance, the performance of the Alpaca-13B with the original instruction templates outperformed its average performance by more than one standard deviation in 7 out of 
10 \lmentry{} tasks.
For lack of space, the figure does not depict the \bbh{} benchmark, but similar patterns of divergence were observed there as well.


In line with \citet{lou2023prompt}, we find that major differences in performance can occur even for very similar paraphrase pairs. 
For example, the \flanlarge{} model demonstrated an average performance degradation of 28\% when changing the word `excludes' to `lacks', while the \flanxl{} model showed an average performance improvement of 46\% on that same edit. 
{See a comprehensive edit distance comparison in Figure~\ref{fig:edit_dist_pairs} and Table~\ref{tab:min_edit_dist_max_diff}.} 

\begin{figure}[t!]
\includegraphics[width=\columnwidth]{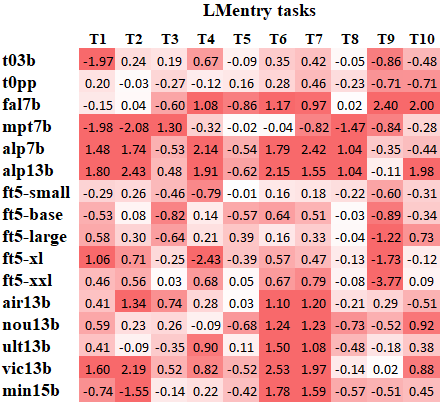}
\caption{\label{fig:divergence_lmentry_bbh}
{Model and task performance divergence. For each \lmentry{} task, we show the number of standard deviations by which  performance of each model on the original instructions deviates from averaged performance. Dark cells indicate substantial divergence values (>1 std). 
}
}
\end{figure}

\begin{figure*}[!htbp]
\includegraphics[width=\textwidth]{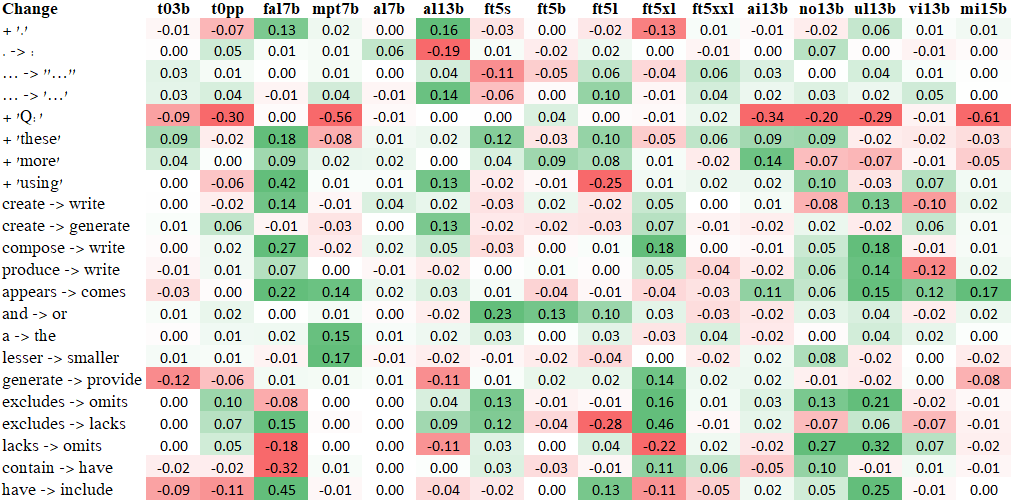}
\caption{\label{fig:edit_dist_pairs}
{Average performance differences between various models for the most common minimal edits between two instruction templates (e.g., substituting `excludes' with `lacks') in the \lmentry{} benchmark.} 
}\end{figure*}

\subsection{LLMs are also Sensitive to Manual Paraphrases}
\label{sec:manual}

Inconsistencies observed in our analyses could stem from paraphrases that leaked to the training of the models. To address this, we extended our analysis with instruction paraphrases which were recently written by~\citet{sun2023evaluating} for the BBL tasks
(7-12 instruction templates per task). 
Importantly, these human-crafted paraphrases were written \emph{after} model training. 

We use 
these annotations to examine model performance.
Our analysis revealed similar inconsistencies as observed with automated paraphrases{, demonstrating model sensitivity to  paraphrasing even when the potential for instruction leakage is minimized.}
{See Table~\ref{tab:tasks_info2} for the Kendall's W values for all BBL tasks, and Figure~\ref{fig:tau-comparison} for a pair of instruction templates exhibiting the minimal Kendall's $\tau$ correlation across all BBL tasks.}


\begin{table*}[!tb]{
\centering
\small
\resizebox{\textwidth}{!}{%
\begin{tabular}{ccp{4.6cm}cp{4.6cm}cc}
\toprule
\textbf{Change}                 & \textbf{Model} & \textbf{P1}                                                                     & \textbf{Acc.} & \textbf{P2}                                                                          & \textbf{Acc.} & \textbf{Diff.}                \\ \midrule
{`.' --> `:'}                                     & nous-hermes          & Create a word that does not include the letter ``\{letter\}''\textbf{\textit{.}}            & .04              & Create a word that does not include the letter ``\{letter\}''\textbf{\textit{:}}             & .65             & {\color[HTML]{00B050} +.61} \\ [15pt]
               & alpaca-13b          & Create a sentence that concludes with the term ``\{word\}''\textbf{\textit{.}}            & .61             & Create a sentence that concludes with the term ``\{word\}''\textbf{\textit{:}}             & .19             & {\color[HTML]{C00000} -.42} \\ [15pt] \midrule
               
               {+ `.'}                      & alpaca-13b          & Write a word that lacks the letter ``{letter}''             & .04             & Write a word that lacks the letter ``{letter}''\textbf{\textit{.}}              & .42             & {\color[HTML]{00B050} +.38} \\ [15pt]
                                     & flan-t5-xl          & Write a word that omits the letter ``{letter}'' & .77             & Write a word that omits the letter ``{letter}''\textbf{\textit{.}} & .54             & {\color[HTML]{C00000} -.23} \\ [15pt] \midrule

                                    {+ `using'} & flan-t5-large           & Your task is to write a word without the letter ``\{letter\}'' & .46             & Your task is to write a word without \textbf{\textit{using}} the letter ``\{letter\}''              & .12             & {\color[HTML]{C00000} -.35} \\ [15pt]
                                     & falcon-7b           & Write a word without the letter \{letter\}.\textbackslash{}nOutput word:             & .12             & Write a word without \textbf{\textit{using}} the letter \{letter\}.\textbackslash{}nOutput word: & .35              & {\color[HTML]{00B050} +.23} \\ [15pt]\midrule
{omits --> lacks}           & ultralm-13b          & Write a word that \textbf{\textit{omits}} the letter ``\{letter\}''.              & .62             & Write a word that \textbf{\textit{lacks}} the letter ``\{letter\}''.                      & .19             & {\color[HTML]{C00000} -.42} \\ [15pt]
                                     & flan-t5-xl          & Write a word that \textbf{\textit{omits}} the letter ``\{letter\}''.                 & .54             & Write a word that \textbf{\textit{lacks}} the letter ``\{letter\}''.                   & .81              & {\color[HTML]{00B050} +.27} \\ [15pt]\midrule
{contain --> have}           & falcon-7b          & Write a word that does not \textbf{\textit{contain}} the letter ``\{letter\}''                 & .81             & Write a word that does not \textbf{\textit{have}} the letter ``\{letter\}''                     & .19             & {\color[HTML]{C00000} -.62} \\ [15pt]
                                     & flan-t5-xxl          & Please write a word that does not \textbf{\textit{contain}} the letter ``\{letter\}''. & .62             & Please write a word that does not \textbf{\textit{have}} the letter ``\{letter\}''.                     & .88              & {\color[HTML]{00B050} +.27} \\ [15pt] \midrule
{include --> have}          & falcon-7b          & Write a word that does not \textbf{\textit{include}} the letter ``\{letter\}''.     & .81             & Write a word that does not \textbf{\textit{have}} the letter ``\{letter\}''.            & .19             & {\color[HTML]{C00000} -.62} \\ [15pt]
& flan-t5-xl & Write a word that does not \textbf{\textit{include}} the letter ``\{letter\}''. & .42 & Write a word that does not \textbf{\textit{have}} the letter ``\{letter\}''. & .73 & {\color[HTML]{00B050} +.31} \\ [15pt]
& ultralm-13b & Please write a word that does not \textbf{\textit{include}} the letter ``\{letter\}''. & .46 & Please write a word that does not \textbf{\textit{have}} the letter ``\{letter\}''. & .12 & {\color[HTML]{C00000} -.35} \\ [15pt] \midrule
{excludes --> lacks}    & flan-t5-large & Write a word that \textbf{\textit{excludes}} the letter ``\{letter\}''.          & .54 & Write a word that \textbf{\textit{lacks}} the letter ``\{letter\}''. & .12 & {\color[HTML]{C00000} -.42} \\ [15pt]
                                     & flan-t5-xl         & Write a word that \textbf{\textit{excludes}} the letter ``\{letter\}''.          & .19 & Write a word that \textbf{\textit{lacks}} the letter ``\{letter\}''. & .81 & {\color[HTML]{00B050} +.62} \\ \bottomrule

\end{tabular}
}
\caption{Representative examples of instruction template pairs from \lmentry{} with very minor differences but notable variations in performance (open-source models).}\label{tab:min_edit_dist_max_diff}
}
\end{table*}

\remove{To estimate this variance, we perform an analysis of variance (one-way ANOVA) F test: $\mathbb{R}^{m \times n} \mapsto \mathbb{R}^{+}$.
The ANOVA tests the null hypothesis, which states that samples in all groups are drawn from populations with the same mean values. In our case, each group is defined by a instruction template- the collection of scores from different LLMs for an instruction template. \rotem{TODO: write updated results from tests here and fix explanation and example} 

This outcome, however, can be explained by the noise generated by the instruction templategeneration process. To refute this reasoning, we first manually classify the generated instruction templates to verify that they are indeed paraphrases of the task instruction \rotem{add results from Moran's analysis and explain our conclusion}. 

Second, we take a closer look at the ANOVA test statistic values. ANOVA decomposes the total variance in the dependent variable into two components: variance between groups and variance within groups.
ANOVA computes the ratio of variance between the means of the observed performances for each instruction templatewith the variance of observed performance across all instruction templates. 

A large ANOVA F-statistic suggests substantial variations in the means of absolute model performance between instruction templateparaphrases for the task, which are unlikely to have occurred by random chance alone. Conversely, a small F-statistic (smaller than 1 and close to 0) indicates that the means of the groups are relatively similar, and any observed differences are more likely to be attributed to variability between the samples. \gabis{@moran and @rotem, please verify}}

\remove{
\paragraph{Sensitivity to Slight instruction templateAdjustments:}
We examined how small changes in instruction templates affected model results. For each model, we looked at template pairs with very minor differences but with notable variations in performance.
Interestingly, every model exhibited instances where marginal instruction templatemodifications led to marked performance declines.Some representative examples of this behavior within the OpenAI model family, when evaluated on the LMentry benchmark, are showcased in Table X.}

\remove{
\paragraph{Robustness to Different Instruction Template Features:}
In evaluating model robustness with respect to instruction templates, we took into account several dimensions: (1) generation method (encompassing rephrasing, chain-of-thought, and progressive template generation); (2) template format (including Q\&A, separated variables, and other); (3) language style; and (4) template length. For the generation method, we ranked all instruction templates by their performance for each model and tabulated the distribution of templates from each method across quartiles (refer to Tables x, y, z in the Appendix)\mnote{TODO: add}. We subsequently compared the distribution in the top quartile to the overall distribution of templates by method. \mnote{TODO: Briefly summarize results here}. We employed a similar methodology for the second and third dimensions (see Tables x, y, z in the Appendix)\mnote{TODO: add}. \mnote{TODO: Briefly summarize results}.

For the evaluation related to template length, we computed Kendall's tau rank correlations between instruction template performances and lengths (Table X in the Appendix)\mnote{TODO: add}. \mnote{TODO: Briefly summarize results}.}

\remove{
\subsection{Quantifying ranking robustness} 
We are interested in observing the importance of instruction variance to reporting proper and comprehensive model capabilities. That is, we want to quantify: \emph{how sensitive are instruction model rankings to the variance in task instructions?}
To quantify how much instability the instruction variance introduces to model rankings we define the following two metrics, that are calculate w.r.t some score function \(S(m,T,I)\):

\paragraph{Hard-Ranking stability}
This metric is defined as follows for a given task test-set \(T\), a set of models \(M\), an instruction distribution \(P(I)\), and a given target model \(m'\):

\begin{equation}
\label{eq:hard_ranking_stability}
    \begin{split}
    R_{\text{hard}}(m' |M, T, P(I))=\\
    \sum_{I \sim P(I)}\left[\sum_{m \sim M}\left[\mathbb{1}_{S(m',T,I)>S(m,T,I)}\right] \right]
    \end{split}
\end{equation}

E.g, given the following model rankings for a task with two instruction templatevariants: a<b<c, c<a<b. 'a' outperformed a single model ('c'). 'b' outperformed both 'a' and 'c' with one instruction variant and 'a' in another. 'c' wins over 'a' and 'b' in one instruction variant. Therefore, when normalized, 'a''s score would be 16.7\% (1/6), 'b''s would be 50\% (3/6), and 'c''s would be 33\% (2/6). 

\paragraph{Soft-score stability}
Although the rankings are important, relaying on the rankings alone might over-score some models although the performance gap between them and lower-rank models is small. To incorporate this notion we define a relaxed score aimed to capture those subtle or unsubtle performance differences. For a given task test-set \(T\), a set of models \(M\), an instruction distribution \(P(I)\), and a given target model \(m'\) we define \emph{Soft-score stability} as:

\begin{equation}
\label{eq:soft_ranking_stability}
\begin{split}
R_{\text{soft}}(m' |M, T, P(I))=\\ 
\sum_{I \sim P(I)}\left[\sum_{m \sim M}\left[S(m',T,I)-min_{m \in M}S(m,T,I)\right] \right]
\end{split}
\end{equation}

This score gives higher values to models dominating over the minimal rank, w.r.t to their distance from the minimal ranking model score. In our analysis we use the ranking scores with the scores defined by Equations \ref{eq:score_max}, \ref{eq:score_mean},  \ref{eq:saturation} (and their combinations), as defined in Section \ref{sec:state_of_what_art}.}

\remove{
\subsection{Robustness to task-specific instructions}

Due to budget considerations, we decided to first illustrate the effect of instructions variants by specifically focusing on semantically equivalent instruction templates from our generated pool, which would lead to higher variations in the scores. \dan{TODO: MORAN, ELABORATE.} We then proceed to evaluate the same effect in BBL, by using the human annotations used in \citet{sun2023evaluating} that were solely curated for introducing more variance.
Figures \ref{fig:lmentry_ranks_example}, \ref{fig:lmentry_ranks_example} \dan{TODO:BBH goes here as well} illustrate the phenomena on tasks from different benchmarks using our generative variants, while Figures \ref{fig:bbl_ranks_example} illustrate the same phenomenon for human-written instruction templates. In all cases, it is apparent that all possible score rankings might occur by using a certain instruction template, out of a pool of templates that are semantically equivalent. The variability induced by instruction instruction templates can be further visualized by examining the differences between the average score (Equation \ref{eq:score_max}) and the score achieved with the best-performing instruction template(Equation \ref{eq:score_mean}) for each model, as seen in Figure \ref{fig:lmentry_max_vs_mean_score}.

To get a numeric sense of this effect we run both stability metrics (normalized) as defined by Equations \ref{eq:hard_ranking_stability}, \ref{eq:soft_ranking_stability} on our chosen benchmarks and use their scores to see whether any model dominated certain tasks while taking instruction instruction templatevariance into account. Examining Figures \ref{fig:bbl-hard-ranks-stabiilty}, \ref{fig:bbl-hard-ranks-stabiilty},\ref{fig:bbl-hard-ranks-stabiilty} \dan{placeholder for BBH and LMEntry} shows that while some model have relatively dominated some task, no model dominates more than 20\% of the times, while figures \ref{fig:bbl-soft-ranks-stability},\ref{fig:bbl-soft-ranks-stability},\ref{fig:bbl-soft-ranks-stability} \dan{placeholder for BBH and LMEntry} show similar trends. \dan{We obviously need to wait for more benchmark results.} 
\dan{TODO: ADD STATISTICAL ANALYSIS HERE too}

Our results are consistent across our instruction instruction templates as well as in results on human written instruction templates acquired from other works. This highlights the importance of defining clearer goals w.r.t instructions and reporting more granular results on benchmarks aimed at quantifying ICL with instruction models.

\begin{figure}[!t]
    \centering
    \includegraphics[width=0.9\linewidth]{figures/lmentry_task_ranking_example.jpeg}
    \caption{
        A bar plot comparing 3 models: gpt-3.5-fast, text-davinchi-002, and text-davinchi-003 on a given LMentry task using different instruction templates. We can see that different instruction templates result in all possible rankings between the model scores.
     }
    \label{fig:lmentry_ranks_example}
\end{figure}

\begin{figure}[!t]
    \centering
    \includegraphics[width=0.9\linewidth]{figures/bbl_task_ranking_example.jpeg}
    \caption{
        A bar plot comparing 3 models: T0++, Alpaca-13b and Flan-T5 XXL on a given BBL task using different human written instruction templates accounted for being unobserved during the training by \citet{sun2023evaluating}. We can see that different instruction templates result in all possible rankings between the model scores.
     }
    \label{fig:bbl_ranks_example}
\end{figure}

\begin{figure}[!t]
    \centering
    \includegraphics[width=0.9\linewidth]{figures/LMenrty_max_vs_mean_performance.jpeg}
    \caption{
        The differences in Score \ref{eq:score_max} and Score \ref{eq:score_mean} emphasize the influence of instruction variants on the performance of instruction models on the LMentry dataset.}
    \label{fig:lmentry_max_vs_mean_score}
\end{figure}

\begin{figure}[!t]
    \centering
    \includegraphics[width=0.87\linewidth]{figures/BBL_hard_ranks.png}
    \caption{
        Leaderboard hard-ranking stability for Big-Bench-Lite. 
        Each row relates to a single task in the BBL benchmark. For each model and task, the score represents how much the model dominated the task w.r.t all of its rankings induced by different instruction templates (as defined by Equation\ref{eq:hard_ranking_stability}). We can observe that although T0++ and Flan-T5 XXL, and Flan-T5 XL dominate the leading relatively, no model is leading absolutely by achieving a score of 50\% or higher (the maximal score was 22\%).
     }
    \label{fig:bbl-hard-ranks-stabiilty}
\end{figure}

\begin{figure}[!t]
    \centering
    \includegraphics[width=0.87\linewidth]{figures/BBL_soft_ranks.png}
    \caption{
        Leaderboard soft-ranking stability for Big-Bench-Lite. 
        Each row relates to a single task in the BBL benchmark. The score in each cell is the average distance in performance from the worst measured performance in that task, over different instruction instruction templates. The score takes into account the score difference with the last rank, and not only the discrete rank (as defined by Equation \ref{eq:soft_ranking_stability}). This enables us to take into account the skew between the values in the ranks, differentiating between high ranks with and without significant score differences from the lowest rank (the scores slope). We can observe that many tasks are not dominated by a particular architecture, while others (language identification, logic grid puzzle) have clearer rankings.
     }
    \label{fig:bbl-soft-ranks-stability}
\end{figure}

\subsection{Robustness to style-specific instructions}
\dan{we thought to ablate the results w.r.t vocabulary level, and the style of the example introduction in the instruction template(whether the input values are intertwined inside the instruction or appear immediately after it}

\subsection{Can different task samples benefit from different instructions?}
\dan{If we run multiple instruction templates for each task, we can ask whether different samples "prefer" different instructions, i.e how instruction dependent are results on given samples? This is interesting and we get this for free}

\subsection{How many instruction templates are needed?}
We use our generated instruction templates to ask: how many instruction templates are needed to approximate the instruction distribution? Furthermore, how does this amount change when applied to domain-specific vs. general instruction templates? 
}

\section{Different Use Cases Merit Different Metrics}
\label{sec:desiderata}
We have shown that LLM performance is greatly affected by paraphrasing of instruction templates.
This calls into question current evaluation practices, which typically rely on LLM performance on a single instruction template.
In this section we explore ways to evaluate LLMs using a  
\emph{{diverse set of instruction templates}}.

Most importantly, we argue that the answer should depend on the \emph{purpose of the evaluation}, and that different extrinsic needs should lead to different evaluation metrics, rather than striving for a coarse catch-all metric.
We introduce a set of metrics, each tailored to specific scenarios and realistic user needs. 

\paragraph{Notations.}
In the following, $M$ is a pretrained LLM, 
 $T = \{(x_i, y_i)\}$  denotes an evaluation dataset for $M$, $I_{T}$ is a set of natural language task instruction paraphrases for $T$ (e.g., obtained via automatic paraphrasing), and $\varepsilon(M, T, i)~\in~[0,1]$ denotes the aggregated performance of $M$ on  samples from $T$, using a single instruction template $i \in I_{T}$ according to a standard metric, e.g., accuracy or $F_1$.

\subsection{Maximum Performance Metric -- For Particular Downstream Applications}
 We define the maximum performance (MaxP) of a model $M$ on task $T$ to be the maximum individual instruction template performance this model achieves
across all instruction templates:
\begin{equation*}
    \label{eq:maxp}
MaxP(M, T, I_T) = \max_{i\in I_T}\;\varepsilon(M, T, i)
\end{equation*}

\textbf{\textit{Use case: }}
This metric is useful for developers aiming to integrate an LLM into a specific downstream task and domain (e.g., sentiment analysis in the news domain).
{In such cases, a user input is often embedded within a fixed instruction template.} 
As such, it makes sense to~find the best-performing instruction template for a given model~\citep{wei2021finetuned}. 
To mitigate overfitting, 
{we advise developers to use a new sample set for the task. This ensures the chosen prompt is validated by its ability to maximize performance on these held-out samples irrespective of prior exposure during training.}

\subsection{Average Performance Metric -- For LLM Developers}
We define the average performance (AvgP) of a model $M$ on task $T$ as the mean of the individual instruction template performances over all instruction templates for the task:
\begin{equation*}
\label{eq:avgp}
AvgP(M, T, I_{T}) = \frac{1}{|I_{T}|} \cdot \sum_{i\in I_{T}}\varepsilon(M, T, i)
\end{equation*}

\textbf{\textit{Use case: }}
Average prompt performance is useful for assessing model robustness to paraphrases. We 
believe this should be standard practice for LLM developers when presenting the performance of a new LLM on a range of tasks and prompt paraphrases \citep{le2022bloom}, as it mitigates outliers in performance. 

\remove{
\subsection{Saturation Metric -- For Open-Ended Applications}
The saturation metric aims to measure how close the model's best performance on a task is to its average performance on that task. 
\begin{align*}
\label{eq:sat1}
Sat(M, T, I_T) =  1 - (MaxP - AvgP)
\end{align*}

\textbf{\textit{Use case: }}
This metric is useful when aiming to deploy an LLM in an environment where instructions might vary – perhaps due to user naturally occurring paraphrases in use inputs or localization to different languages. A high saturation score indicates that the model's performance does not drop significantly for non-optimal instructions. Thus, the model is versatile and adaptable to diverse real-world scenarios.

\subsection{Combined Performance Score}
The combined performance score takes into account the model's best performance and its saturation. It serves as a single metric capturing both the peak capability and consistency of the model across prompts. \
\begin{equation*}\label{eq:cps}
CPS(M, T, I_T) = Sat \cdot MaxP
\end{equation*}
\textbf{\textit{Use case: }}
This metric is useful for selecting a model for a suite of applications or a platform offering diverse tasks. Consider a scenario aiming to integrate an LLM into an application in which some functionalities might have user-visible prompts. One example of such an application is a multi-functional chatbot – the chatbot needs to be both effective (high performance) and flexible (consistent across varying instructions). CPS aids in identifying models that strike this balance.
}

\subsection{Combined Performance Score}
In the same way the F1 score combines precision and recall into a single metric, we propose a Combined Performance Score (CPS) that unites the maximum and average performance metrics to capture both peak capability and {robustness} of the model across prompts.
To define CPS, we first introduce a model saturation score:
\begin{align*}
\label{eq:sat1}
Sat(M, T, I_T) = 1 - (MaxP - AvgP)
\end{align*}
This score measures how closely the model's best performance aligns with its average performance. A high saturation score indicates that the model's performance does not drop significantly for non-optimal instructions. Then, the CPS is calculated as the product of the model’s best performance ($MaxP$) and its saturation ($Sat$):
\begin{equation*}\label{eq:cps}
CPS(M, T, I_T) = Sat \cdot MaxP
\end{equation*}

\textbf{\textit{Use case: }}
This metric is valuable for selecting a model for a suite of applications or a platform offering diverse tasks. For instance, when integrating an LLM into an application with user-visible prompts, such as a multi-functional chatbot, it is crucial for the model to be both effective (high $MaxP$) and {robust} (high $Sat$). CPS facilitates identifying models that strike a balance between top-tier performance and {robust} reliability across varying instruction templates.



\remove{
\paragraph{Best Instruction Performance:} This score measures the performance of the model with the prompt that produces the best mean results on the task inputs. This metric is a proxy for measuring the maximum potential performance of the instruction model for a given task and can be expressed as the scoring function \(S_{\text{best}}(M, D)\):

\begin{equation}\label{eq:score_mean}
S_{\text{best}}(m, D) = \mathbb{E}_{T \sim p(T)} \left[ \max_{I \sim p(I)} \left[ s(m, T, I) \right] \right]
\end{equation}

\paragraph{Mean Instruction Performance:} This score calculates the average performance over all prompts given a task sample, thereby capturing the average performance and robustness of the model across human instructions. This metric can be represented by the scoring function $S_{\text{mean}}(m, D)$:

\begin{equation}\label{eq:score_max}
S_{\text{mean}}(m, D) = \mathbb{E}_{T \sim p(T)} \left[ \mathbb{E}_{I \sim p(I)} \left[ s(m, T, I) \right] \right]
\end{equation}

where $s(m, T, I)$ evaluates the performance of the instruction model \(m\) on a specific task sample \(T\) and instruction \(I\).

\paragraph{Relative Performance:} It can also be valuable to report a score that captures how much of the model's maximal potential is achieved on average with different instructions. This metric, denoted as \(S_{\text{potential}}(M, D)\), compares the model's average performance to the best achievable performance on each task, capturing the model's relative performance:
\begin{equation}
\label{eq:saturation}
S_{\text{potential}}(m, D) = \frac{S_{\text{mean}}(m, D)}{S_{\text{best}}(m, D)}
\end{equation}

Note that these metrics and considerations can be further refined and tailored to specific evaluation criteria and objectives of instruction models, using appropriate evaluation functions 
\(s(m, T, I)\) that are aligned with the task and desired performance measures.
}

\section{Multi-Prompt Evaluation}
\label{sec:desiderta_implementation}

In Figure~\ref{fig:cross-evaluation} we evaluate all our \numofmodelsopen{} models according to the metrics we proposed in the previous section, on 
sample tasks from each of the three benchmarks {(full results for all tasks are available in our repository)}. 
We report several interesting observations.
First, we find that all aggregate metrics diverge from the performance on the 
original instruction templates.
For the vast majority of the tasks in our study, the top three models determined by the original instruction templates were different from those which ranked first according to the average and maximum metrics.

More broadly, model ranking depended on the metric used. For instance, see Figure \ref{fig:cross-evaluation} 
 (top): In \lmentry{}'s rhyming word task, \falcon{} and \vicunawp{} rank first according to $MaxP$ (0.74, gray and yellow bars), but their average performances $AvgP$ are only 0.17 and 0.15, respectively. 
Similarly, across all tasks in the \lmentry{} benchmark, \llama{}-based models were competitive with T5-based models in terms of $MaxP$. However, in terms of $AvgP$, they tended to lag behind, due to extremely poor performance on a large number of paraphrases (see Figure~\ref{fig:notable_failures_lmentry} for \%paraphrases that achieved at least $5\%$ accuracy). 

Finally, we found that noise stemming from automatic paraphrase generation has virtually no impact on metric-based model rankings.
We compute Kendall's $\tau$ to compare model rankings before and after the manual filtering of paraphrases. The results (Table~\ref{tab:metric_rankings_comparisson_avg}) show near-perfect to perfect agreement in rankings across all tasks, except for the “ends with word” task in \lmentry{}. Upon examination, this seems to be mostly due to an error in \lmentry{}'s evaluation script.
These results suggest that it may be enough to compute our metrics over range of automatically-generated paraphrases, without having to manually verify them.

\begin{figure}[t!]
\includegraphics[width=\linewidth]{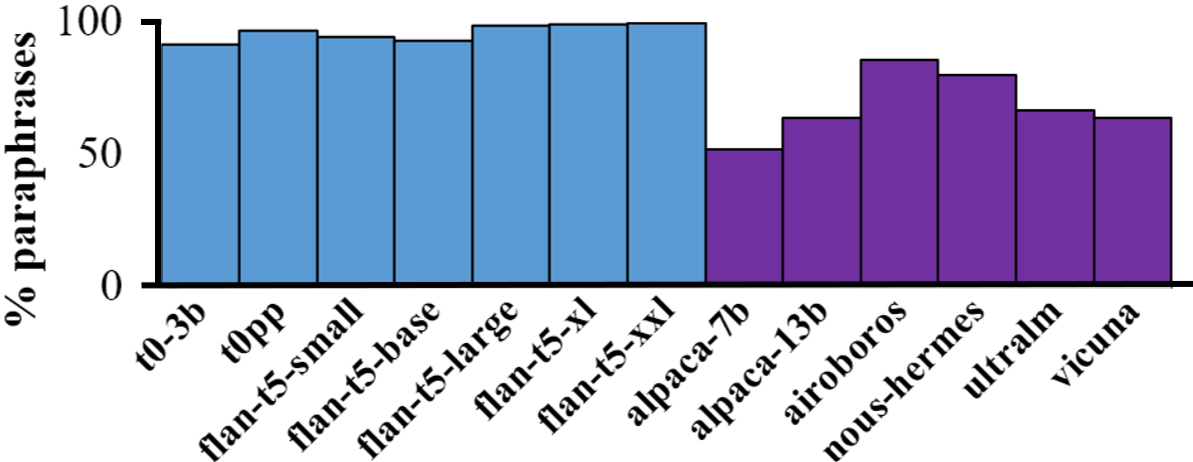}
\caption{\label{fig:notable_failures_lmentry}
Percentage of instruction paraphrases with accuracy higher than 5\% in T5 models (blue) vs. \llama{} models (purple) on \lmentry{} tasks.
}\end{figure}

\begin{table}[t!]
\centering
\renewcommand{\arraystretch}{1.3}
\LARGE
\resizebox{\columnwidth}{!}{%
\begin{tabular}{@{}lccc@{}}
\toprule
\textbf{Benchmark}& \textbf{Max perf.} & \textbf{Average perf.} & \textbf{Combined perf.}
\\ \midrule
\lmentry{}  & .963 & .978  & .948\\
\bbh{}  & .991 & .983  & .966\\ \bottomrule
\end{tabular}%
}
\caption{Averaged Kendall’s Tau values comparing rankings before and after filtering incorrect paraphrases for each metric across all tasks (excluding ``ends with word'' for \lmentry{}). 
}
\label{tab:metric_rankings_comparisson_avg}
\end{table}

\begin{figure}[t!]
\center
\includegraphics[width=\linewidth]{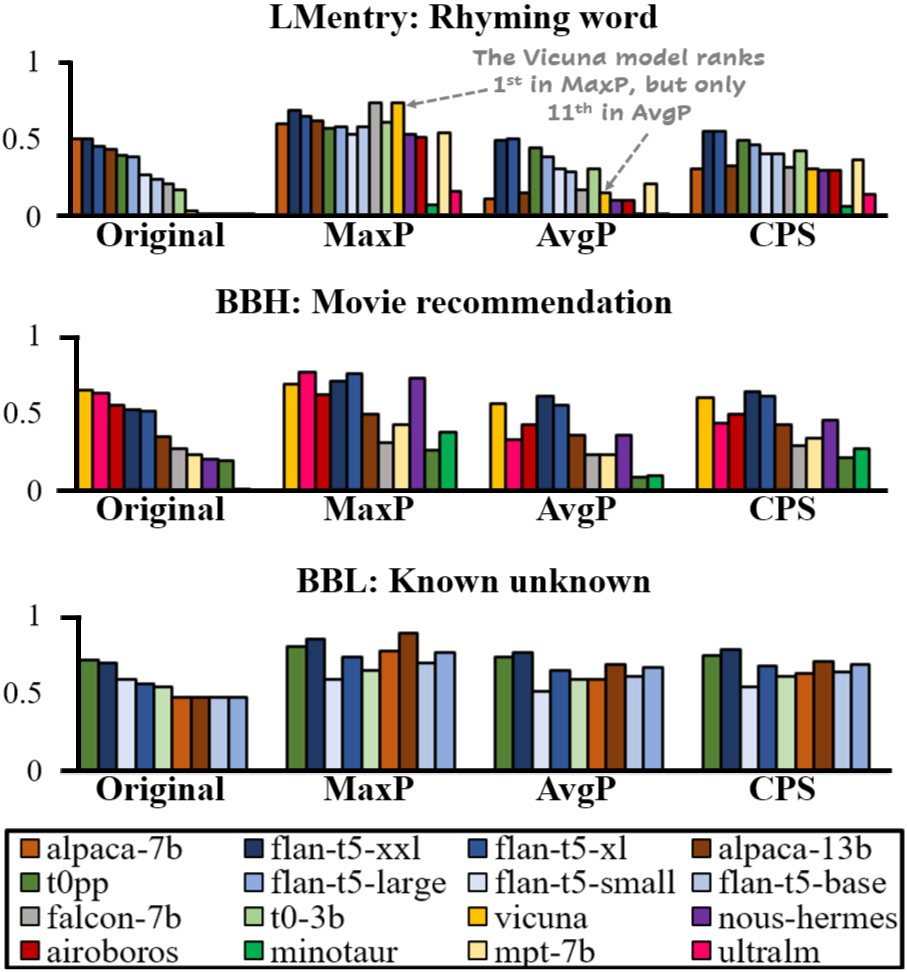}
\caption{\label{fig:cross-evaluation} The performance of various models according to the metrics proposed in Section \ref{sec:desiderata}, evaluated on sample tasks from each of the three benchmarks. The name of the metric appears below each group of columns; height of a column represents value in \emph{that specific metric}. The order of the columns (i.e., models) between groups is fixed, set according to decreasing performance on the original instruction templates {to enable straightforward comparisons of ranking changes}.  
}
\end{figure}

\section{Small-Scale Evaluation of OpenAI Models on Prompt Paraphrasing}
\label{sec:openai_evaluation}

In this section we perform a small-scale evaluation showing that API LLMs are also sensitive to instruction paraphrasing. Our evaluation focuses on four OpenAI models: davinci, text-davinci-002, text-davinci-003, and \gptturbo{} on the \lmentry{} benchmark. 

Due to budget constraints, we show that the performance of these models diverges significantly between the benchmark's original instruction templates and a selection of paraphrases, in terms of both  average and maximum metrics.


\begin{table*}[t!]{
 \resizebox{\textwidth}{!}{%
\centering
\small
\begin{tabular}{ccp{5.5cm}cp{5.5cm}cc}
\toprule
\textbf{Change}                 & \textbf{Model} & \textbf{P1}                                                                     & \textbf{Acc.} & \textbf{P2}                                                                          & \textbf{Acc.} & \textbf{Diff.}                \\ \midrule
{\{...\} --> ``\{...\}''}                                     & td002          & Which word has a  greater number of letters, \textbf{\textit{\{word1\}}} or \textbf{\textit{\{word2\}}}?            & .50              & Which word has a  greater number of letters, \textbf{\textit{``\{word1\}''}} or \textbf{\textit{``\{word2\}''}}?             & .23             & {\color[HTML]{C00000} -0.27} \\ [15pt]
               & td002          & Which of the words \textbf{\textit{\{word1\}}} and \textbf{\textit{\{word2\}}}  is alphabetically first?            & .54             & Which of the words \textbf{\textit{``\{word1\}''}} and \textbf{\textit{``\{word2\}''}}  is alphabetically first?             & .77             & {\color[HTML]{00B050} +0.23} \\ [15pt]
                                     & td003          & Which word has a greater number of letters, \textbf{\textit{\{word1\}}} or \textbf{\textit{\{word2\}}}?             & .60             & Which word has a greater number of letters, \textbf{\textit{``\{word1\}''}} or \textbf{\textit{``\{word2\}''}}?              & .14             & {\color[HTML]{C00000} -0.46} \\ [15pt]
                                     & td003          & Compare the length of \textbf{\textit{\{word1\}}} and \textbf{\textit{\{word2\}}} and tell me which one is shorter. & .39             & Compare the length of \textbf{\textit{``\{word1\}''}} and \textbf{\textit{``\{word2\}''}}  and tell me which one is shorter. & .73             & {\color[HTML]{00B050} +0.34} \\ [15pt]
                                     
                                     & cgpt           & Which word has a greater number of letters, \textbf{\textit{\{word1\}}} or \textbf{\textit{\{word2\}}}?             & .55             & Which word has a greater number of letters, \textbf{\textit{``\{word1\}''}} or \textbf{\textit{``\{word2\}''}}?              & .24             & {\color[HTML]{C00000} -0.31} \\ [15pt]
                                     & cgpt           & Compare the length of \textbf{\textit{\{word1\}}} and \textbf{\textit{\{word2\}}}. Which one is longer?             & .04             & Compare the length of \textbf{\textit{``\{word1\}''}} and \textbf{\textit{``\{word2\}''}}. Which one is longer?              & .70              & {\color[HTML]{00B050} +0.66} \\ [15pt]\midrule
{`,' --> `:'}           & td002          & Which word is a rhyme for ``\{query\}''\textbf{\textit{,}} ``\{word1\}'' or ``\{word2\}''?              & .08             & Which word is a rhyme for ``\{query\}''\textbf{\textit{:}} ``\{word1\}'' or ``\{word2\}''?                   & .85             & {\color[HTML]{00B050} +0.77} \\ [15pt]
                                     & td003          & Which word is a rhyme for ``\{query\}''\textbf{\textit{,}} ``\{word1\}'' or ``\{word2\}''?              & .48             & Which word is a rhyme for ``\{query\}''\textbf{\textit{:}} ``\{word1\}'' or ``\{word2\}''?                   & .90              & {\color[HTML]{00B050} +0.42} \\ [15pt]\midrule
{`,' --> `-'}           & td002          & Which word rhymes with ``\{query\}''\textbf{\textit{,}} ``\{word1\}'' or ``\{word2\}''?                 & .06             & Which word rhymes with ``\{query\}'' \textbf{\textit{-}} ``\{word1\}'' or ``\{word2\}''?                     & .73             & {\color[HTML]{00B050} +0.67} \\ [15pt]
                                     & td003          & Which word rhymes with ``\{query\}''\textbf{\textit{,}} ``\{word1\}'' or ``\{word2\}''?                 & .17             & Which word rhymes with ``\{query\}'' \textbf{\textit{-}} ``\{word1\}'' or ``\{word2\}''?                     & .60              & {\color[HTML]{00B050} +0.43} \\ [15pt] \midrule
{the --> a}          & td002          & What is \textbf{\textit{the}} word that rhymes with ``\{query\}'' - ``\{word1\}'' or ``\{word2\}''?     & .03             & What is \textbf{\textit{a}} word that rhymes with ``\{query\}'' - ``\{word1\}'' or ``\{word2\}''?            & .78             & {\color[HTML]{00B050} +0.75} \\ [15pt]\midrule
{which --> what}    & td002          & \textit{\textbf{Which}} word rhymes with ``\{query\}'' - ``\{word1\}'' or ``\{word2\}''?                & .73             & \textbf{\textit{What}} word rhymes with ``\{query\}'' - ``\{word1\}'' or ``\{word2\}''?                      & .82             & {\color[HTML]{00B050} +0.09} \\ [15pt]
                                     & td003          & \textbf{\textit{Which}} word rhymes with ``\{query\}'' - ``\{word1\}'' or ``\{word2\}''?                & .60              & \textbf{\textit{What}} word rhymes with ``\{query\}'' - ``\{word1\}'' or ``\{word2\}''?                      & .15             & {\color[HTML]{C00000} -0.45} \\ [15pt]\midrule

{word --> term}     & td002          & Create a \textbf{\textit{word}} that excludes the letter ``\{letter\}''.                            & .54             & Create a \textbf{\textit{term}} that excludes the letter ``\{letter\}''.                                 & .04             & {\color[HTML]{C00000} -0.50} \\ [15pt]
                                     & td003          & Create a \textbf{\textit{word}} that excludes the letter ``\{letter\}''.                            & .96             & Create a \textbf{\textit{term}} that excludes the letter ``\{letter\}''.                                 & .58             & {\color[HTML]{C00000} -0.38} \\ [15pt]
                                     & cgpt           & Create a \textbf{\textit{word}} that excludes the letter ``\{letter\}''.                            & .81             & Create a \textbf{\textit{term}} that excludes the letter ``\{letter\}''.                                 & .42             & {\color[HTML]{C00000} -0.39} \\ \bottomrule
\end{tabular}%
}
\caption{Minimal distance 
pairs from \lmentry{} with large performance differences in OpenAI models.} 
\label{tab:openai_edit_distance}
}
\end{table*}

\paragraph{Estimating average performance.}
To estimate the average performance of OpenAI models on a specific task, we adopted a randomized approach. For each task sample, we 
randomly 
selected a paraphrase from our collection, and evaluated the model's response, scoring the entire set of task samples. To approximate average performance, this experiment was repeated 20 times, determined by the data from our 16 open-source~models.

\paragraph{Estimating maximal performance.}
To estimate which of the roughly 175 instruction templates per task performs the best for each model, we implemented a simple greedy search. Initially, we evaluated all paraphrases on 10 task instances, then narrowed down to the top 100 instruction templates for another 10 instances. Finally, the top 10 instruction templates were evaluated on the remaining instances, and the  template that performed the best was chosen to estimate the maximum performance.

\subsection{Results}
Below we summarize the results of our evaluation of OpenAI models. The full details appear in 
our repository.\footnotemark[\getrefnumber{note1}]

\paragraph{OpenAI models are also sensitive to minor prompt variations.}
Minor changes in the phrasing of the instruction could lead to drastic performance changes for the OpenAI models, similar to our findings in Section~\ref{sec:brit_results} with smaller-scale LLMs. 
See representative examples in Table~\ref{tab:openai_edit_distance}, showing nearly identical instruction template pairs resulting in notable variations in performance. 

\paragraph{Average performance is lower than that observed in the original benchmark instructions.}
In 72.5\% of the cases, the performance of the original instructions was higher than the estimated average across all paraphrases. In the davinci model, the original prompts added on average 21 more accuracy points. 

\begin{figure}[t!]
\includegraphics[width=\linewidth]{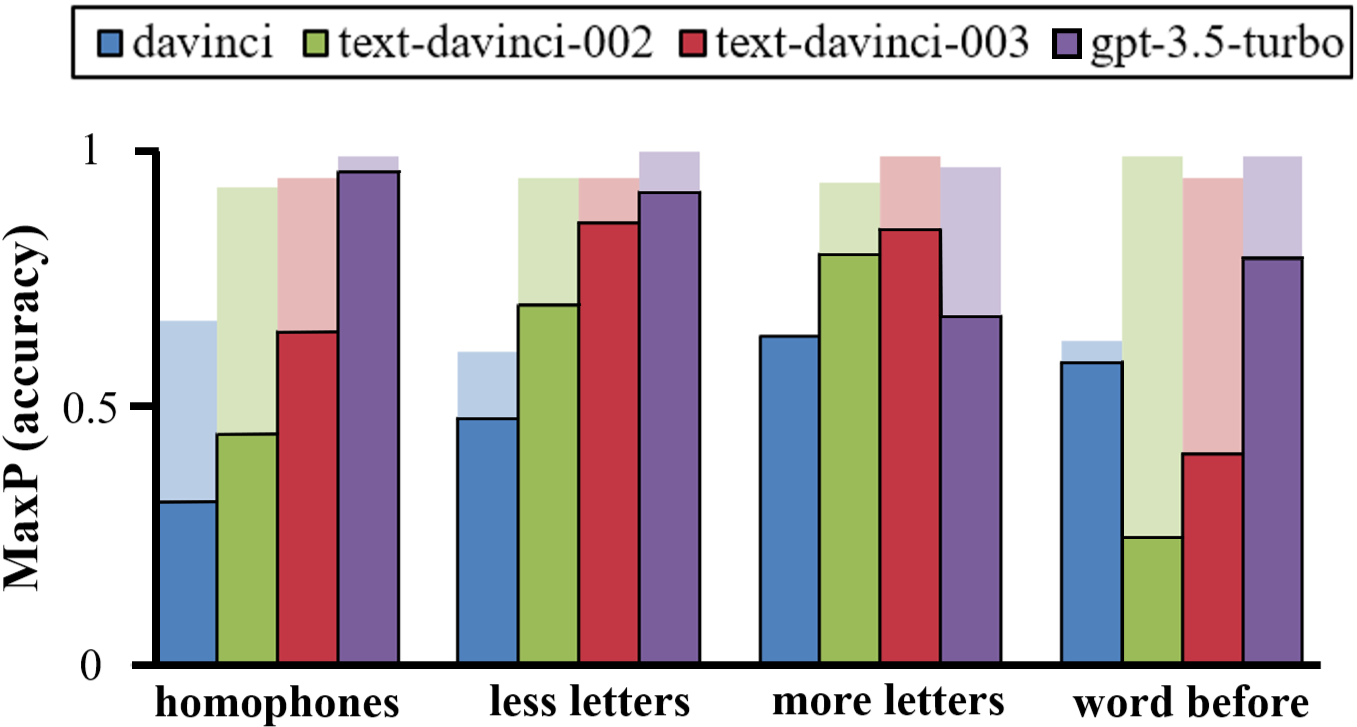}
\caption{\label{fig:max_diff_openai}
Comparison of the \emph{maximum performance} of four OpenAI models using original prompts (in solid colors) vs.~all prompt paraphrases (semi-transparent). Each group of columns corresponds to a different task in the \lmentry{} benchmark. 
}\end{figure}

\paragraph{Original prompt performances fall below all paraphrases’ estimated maximum performance.}
Figure~\ref{fig:max_diff_openai} depicts maximum performance of the \emph{original instructions} for four \lmentry{} tasks in solid colors, with overlaid semi-transparent columns indicating the estimated maximum performance on \emph{all paraphrases}. 
Notably, for  text-davinci-002, we found paraphrases that improved its maximal accuracy performance above 90\% for 8 out of 10 tasks. Across all four models, 26 out of 40 differences were  statistically significant according to the McNemar test.

\paragraph{Model rankings diverge between the different metrics and original instruction templates.}
Similarly to our main evaluation, there were many mismatches between ranking on the original instruction templates and our metrics. Agreement was observed in only 5 out of 10 tasks for the average metric, and in 4 out of 10 tasks for the maximum metric.


\section{Related Work}
\label{sec:related_work}

Our work is part of an emerging trend highlighting the many challenges standing in the way of meaningful, scalable, and reproducible evaluation of large language models. 

\citet{Perlitz2023EfficientB} focus on the rising cost of exhaustive evaluation of LLMs on large number of samples. 
They developed methods for choosing subsets of the test data which are expected to be a good representative of the whole. 
An interesting avenue for future work can extend \citet{Perlitz2023EfficientB}'s approach to also include various instruction templates, thus efficiently approximating our suggested evaluation methods.

\citet{Sclar2023QuantifyingLM} show that LLMs are sensitive to \emph{prompt formatting}. These are minor prompt design choices, such as the addition or omission of punctuation marks. They create a large pool of instruction paraphrases, 
ensuring that paraphrases maintain the meaning of the original prompt. We notice a similar phenomenon,
albeit more anecdotally, when our automatic paraphrasing techniques incidentally produce minor changes in formatting (Table~\ref{tab:openai_edit_distance}). 
%
\citet{voronov2024mind} showed that LLMs are sensitive to the format of in-context examples. For example, they varied the manner in which each input-output is separated, and test how such choices interact with the phrasing of the instruction template, the number of demonstrations, or the model size.

{The works discussed above represent a distinct thread within the larger field of model robustness, which is typically defined as a measure of models' ability to adapt to distribution shifts between training and inference~\citep{wang-etal-2022-measure}, or to cope with adversarial examples~\citep{wang2021adversarial,wang2023robustness}. In contrast, these works do not change the underlying instance to be classified (e.g., the homophone pairs in our running example), but rather the task \emph{instruction}. This challenge arises with the introduction of LLMs which take such instructions as part of the input, rather than through dedicated calibration in training or finetuning.} 

%



\section{Conclusions}
\label{sec:conclusion and Future Work}
Our research highlights the sensitivity of large language models (LLMs) to prompt paraphrasing, challenging the adequacy of single-prompt evaluations. 
We propose alternative evaluation metrics that use a diverse set of instruction templates for each task, designed for more robust and meaningful LLM evaluation. For example, LLM developers may be interested in measuring the robustness of performance across multiple prompts, which we propose to evaluate as the average across a large collection of prompts. In contrast, when developing a downstream model, different models should be compared according to their corresponding top-performing prompt.

Evaluating based on these metrics underscores the necessity for nuanced evaluation methods, revealing notable differences in absolute performance and relative model rankings compared to traditional evaluations. 
We hope that our work will help spur more consistency and comparability in LLM evaluation
which is strongly coupled to real-world LLM uses.
We believe this shift is crucial for accurately understanding and leveraging the true capabilities of LLMs.

\section*{Acknowledgements}
We thank the reviewers for their insightful comments. We further thank Asaf Yehudai and Oyvind Tafjord
 for engaging discussions, and the members of \textbf{\href{https://gabrielstanovsky.github.io/group/}{SLAB}} and \textbf{\href{https://www.hyadatalab.com/}{Hyadata Lab}} at the Hebrew University of Jerusalem for their thoughtful remarks. 
This work was supported by the European Research Council (ERC) under the European Union's Horizon 2020 research and innovation programme (grant no. 852686, SIAM) and was partially supported by the Israeli Ministry of Science and Technology (grant no. 2336).


\bibliography{tacl2021}
\bibliographystyle{acl_natbib}
\clearpage


\end{document}